\documentclass{article}

\usepackage{times}
\usepackage{graphicx} 
\usepackage{subfigure} 

\usepackage{natbib}

\usepackage{algorithm}
\usepackage{algorithmic}

\usepackage[accepted,nohyperref]{icml2017}

\icmltitlerunning{Prior-aware Dual Decomposition (PADD): Document-specific Topic Inference for Spectral Topic Models}

\usepackage{url}            
\usepackage{booktabs}       
\usepackage{amsfonts}       
\usepackage{nicefrac}       
\usepackage{microtype}      

\usepackage{hyphenat}
\usepackage{amsmath}
\usepackage{amssymb}
\usepackage{bm}
\usepackage{multirow}
\usepackage{units}
\usepackage{wrapfig}

\newcommand{\In}{\mathbin{\in}}
\newcommand{\Eq}{\mathbin{=}}
\newcommand{\Ceq}{\mathbin{:=}}

\newcommand{\Plus}{\mathbin{+}}
\newcommand{\Minus}{\mathbin{-}}
\newcommand{\Times}{\mathbin{\times}}

\DeclareMathOperator*{\argmax}{arg\,max}

\newcommand{\Lagr}{\mathcal{L}}

\newcommand{\0}{\bm{0}}

\newcommand{\Bb}{\bm{b}}

\newcommand{\Bf}{\bm{f}}
\newcommand{\Bh}{\bm{h}}
\newcommand{\Bhtilde}{\smash{\widetilde{\bm{h}}}}

\newcommand{\Bp}{\bm{p}}
\newcommand{\Bq}{\bm{q}}

\newcommand{\Bw}{\bm{w}}
\newcommand{\Bwbar}{\bar{\bm{w}}}
\newcommand{\BA}{\bm{A}}

\newcommand{\BB}{\bm{B}}

\newcommand{\BBinv}{\bm{B}^\dagger}
\newcommand{\BBreve}{\smash{\bm{\breve{B}}}}
\newcommand{\BC}{\bm{C}}

\newcommand{\BF}{\bm{F}}
\newcommand{\BG}{\bm{G}}
\newcommand{\BH}{\bm{H}}
\newcommand{\BI}{\bm{I}}

\newcommand{\BHtilde}{\smash{\widetilde{\bm{H}}}}

\newcommand{\BW}{\bm{W}}

\newcommand{\Balpha}{\bm{\alpha}}
\newcommand{\Bbeta}{\bm{\beta}}
\newcommand{\BLambda}{\bm{\Lambda}}
\newcommand{\Bmu}{\bm{\mu}}
\newcommand{\BSigma}{\bm{\Sigma}}

\newcommand{\f}{\mathfrak{f}}
\newcommand{\g}{\mathfrak{g}}

\newcommand{\LN}{\mathcal{LN}}
\newcommand{\R}{\mathbb{R}}

\newcommand{\E}{\mathbb{E}}

\newcommand{\Dir}{\mathrm{Dir}}
\newcommand{\Mult}{\mathrm{Mult}}

\begin{document} 

\twocolumn[
\icmltitle{Prior-aware Dual Decomposition: \\Document-specific Topic Inference for Spectral Topic Models}




\begin{icmlauthorlist}
\icmlauthor{Moontae Lee}{cs}
\icmlauthor{David Bindel}{cs}
\icmlauthor{David Mimno}{is}
\end{icmlauthorlist}

\icmlaffiliation{cs}{Cornell University, Department of Computer Science, USA}
\icmlaffiliation{is}{Cornell University, Department of Information Science, USA}

\icmlcorrespondingauthor{Moontae Lee}{moontae@cs.cornell.edu}

\icmlkeywords{boring formatting information, machine learning, ICML}

\vskip 0.3in
]



\printAffiliationsAndNotice{}  

\begin{abstract} 
Spectral topic modeling algorithms operate on matrices/tensors of word co-occurrence statistics to learn topic-specific word distributions.
This approach removes the dependence on the original documents and produces substantial gains in efficiency and provable topic inference, but at a cost: the model can no longer provide information about the topic composition of individual documents.
Recently Thresholded Linear Inverse (TLI) is proposed to map the observed words of each document back to its topic composition.
However, its linear characteristics limit the inference quality without considering the important prior information over topics.
In this paper, we evaluate Simple Probabilistic Inverse (SPI) method and novel Prior-aware Dual Decomposition (PADD) that is capable of learning document-specific topic compositions in parallel.
Experiments show that PADD successfully leverages topic correlations as a prior, notably outperforming TLI and learning quality topic compositions comparable to Gibbs sampling on various data.

\end{abstract} 

\section{Introduction}
Unsupervised topic modeling represents a collection of documents as a combination of \textbf{topics}, which are distributions over words, and document \textbf{compositions}, which are distributions over topics \cite{Hof,LDA}.
Though topic modeling is commonly applied to text documents, because of no assumptions about word order or syntax/grammar, topic models are flexibly applicable to any data that consists of groups of discrete observations, such as movies in review datasets and songs in playlists in collaborative filtering \cite{moontae2015nips}, or people present in various events in social network \cite{liu2009}.
Topics are useful for quickly assessing the main themes, common genres, and underlying communities in various types of data.
But most applications of topic models depend heavily on document compositions as a way to retrieve documents that are representative of query topics or to measure connections between topics and metadata like time variables \cite{BL1,steyvers2008rational,hall2008studying,talley2011database,goldstone-14,erlin2017topic}.
Learning topic compositions is particularly useful because one can compactly summarize individual documents in terms of topics rather than words.\footnote{The number of topics is by far smaller than the size of vocabulary in the general \textit{non-overcomplete} settings.}

Spectral topic models have emerged as alternatives to traditional likelihood-based inference such as Variational Bayes \cite{LDA} or Gibbs Sampling \cite{griffithsFinding}.
Spectral methods do not operate on the original documents, but rather on summary statistics.
Once constructing the higher-order moments of word co-occurrence as statistically unbiased estimators for the generative process, we can perform moment-matching via matrix or tensor factorization.
The Anchor Word algorithms \cite{AGM,arora2013practical,bansal2014,moontae2015nips,huang2016} factorize the second-order co-occurrence matrix between pairs of words in order to match its posterior moments.
Tensor decomposition algorithms \cite{anandkumar2012,anandkumar2012b,anandkumar2013} factorize the third-order tensor among triples of words toward matching its population moments.

Comparing to the traditional inference algorithms, these spectral methods have three main advantages.
First, training is transparent and deterministic. We can state exactly why the algorithms make each decision, and the results do not change with random initializations.
Second, we can make provable guarantees of optimality given reasonable assumptions, such as the existence of topic-specific \textit{anchor words} in matrix models or uncorrelated/sparse topics in tensor models.
Third, because the input to the algorithm is purely in terms of word-word relationships, we can limit our interaction with the training documents to a single trivially-parallelizable pre-processing step to construct the co-occurrence statistics.
Then we can learn various sizes of topic models without revisiting the training documents.

But the efficiency advantage of factoring out the documents is also a weakness: we lose the ability to say anything about the documents themselves.
In practice, users of spectral topic models must go back and apply traditional inference on the original documents. 
That is, given topic-word distributions and a sequence of words for each document, they need to estimate the posterior probability of topics, still with the assumption of a sparse Dirichlet prior on the topic composition.
Estimating topics with a Dirichlet prior is NP-Hard even for trivial models \cite{SR}.
Long running of Gibbs Sampling for topic inference can be relatively accurate, but has no provable guarantees \cite{yao2009efficient}.
Using Varitional Bayes often falls in local-minima, learning inconsistent models for various numbers of topics.

Recently, \cite{arorab16} proposes the Thresholded Linear Inverse (TLI) method to learn the document-specific topic distributions for spectral topic models.
TLI tries to compute one single inverse of word-topic matrix so that it can inversely transform the term-frequency vector of each document back to a topic composition vector.
Unfortunately the matrix inversion is expansive for large vocabularies, and numerically unstable, often producing NaN entries and thereby learning inferior compositions to likelihood-based inference.
Even though TLI is armed with the provable guarantees, its thresholding scheme turns out to quickly lose both precision and recall as the number of topics increases. 
Due to the lack of prior knowledge over topics in inference, TLI proposes unlikely combinations of topics, being further degraded when the given document is not long enough to provide sufficient information.

\begin{figure}[t]
 	\centering
 		\includegraphics[width=0.45\textwidth, trim={0.0cm 0.0cm 0.0cm 0.0cm}]{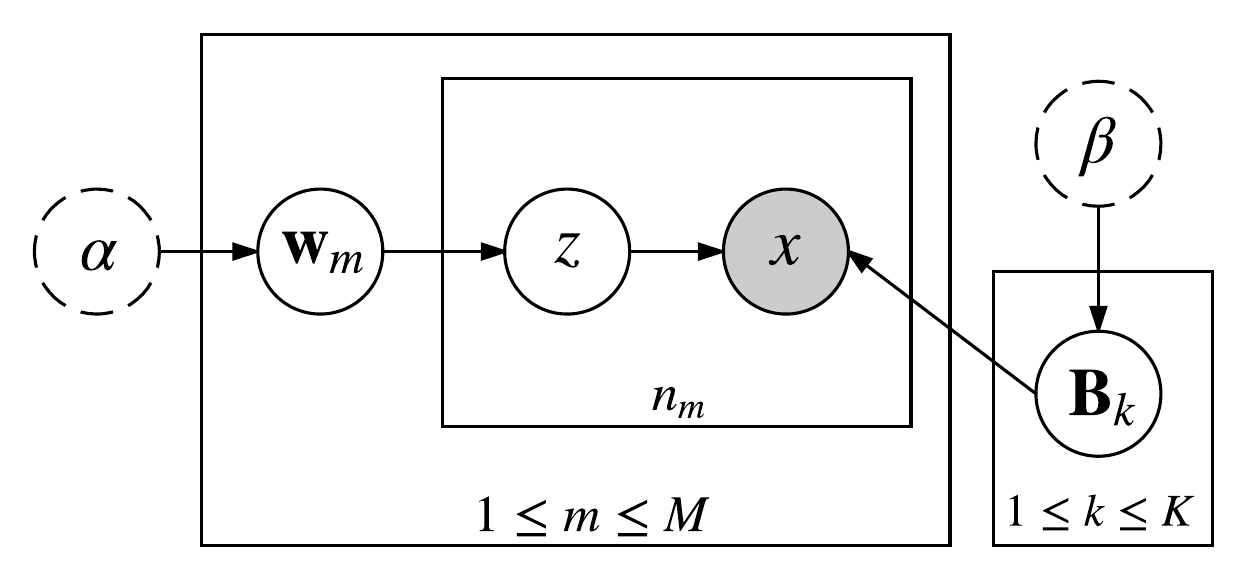}
		\caption {\small{LDA asserts a topic composition $\Bw_m$ for each document $m$. $\Dir(\Balpha)$ provides prior information for the entire corpus.}}
		\label{fig:LDA}	
		\vspace{-15px}
\end{figure}

In this work we propose one benchmark method and another elaborated method.
They perform optimally in different settings, in which the critical difference is the degree of correlation between topics. 
If topics appear independently, with no meaningful correlation structure, a \textbf{Simple Probabilistic Inverse (SPI)} performs well.
In contrast to TLI, SPI uses word-specific topic distribution, which
is the natural and probabilistically coherent linear estimator.
If there are complex correlations between topics, as is often the case in real data, our \textbf{Prior-aware Dual Decomposition (PADD)} is capable of learning quality document-specific topic compositions by leveraging the learned prior over topics in terms of topic correlations.
PADD regularizes topic correlations of each document to be not too far from the overall topic correlations, thereby guessing reasonable compositions even if the given document is not sufficiently long.

Since second-order spectral models naturally learn topic correlations from the decomposition of co-occurrence statistics, this paper focuses mostly on the framework of \textbf{Joint Stochastic Matrix Factorization (JSMF)}.
The rectified anchor word algorithm in JSMF is proven to learn quality topics and their correlations comparable to probabilistic inference \cite{moontae2015nips}.
Above all, it is shown that most large topic models satisfy the anchor word assumption \cite{ding2015}.
However, third-order tensor models can also use our PADD by constructing the moments from some parametric priors that are capable of modeling topic correlations \cite{arabshahi2016spectral}. 
In that case, the second-order population moment estimated by their learned hyper-parameters serves as a surrogate for the prior information over topics, allowing PADD to infer topic compositions.

\section{Foundations and Related Work}

\begin{figure}[t]
	\centering
		\includegraphics[width=0.45\textwidth, trim={0.0cm 0.0cm 0.0cm 0.0cm}]{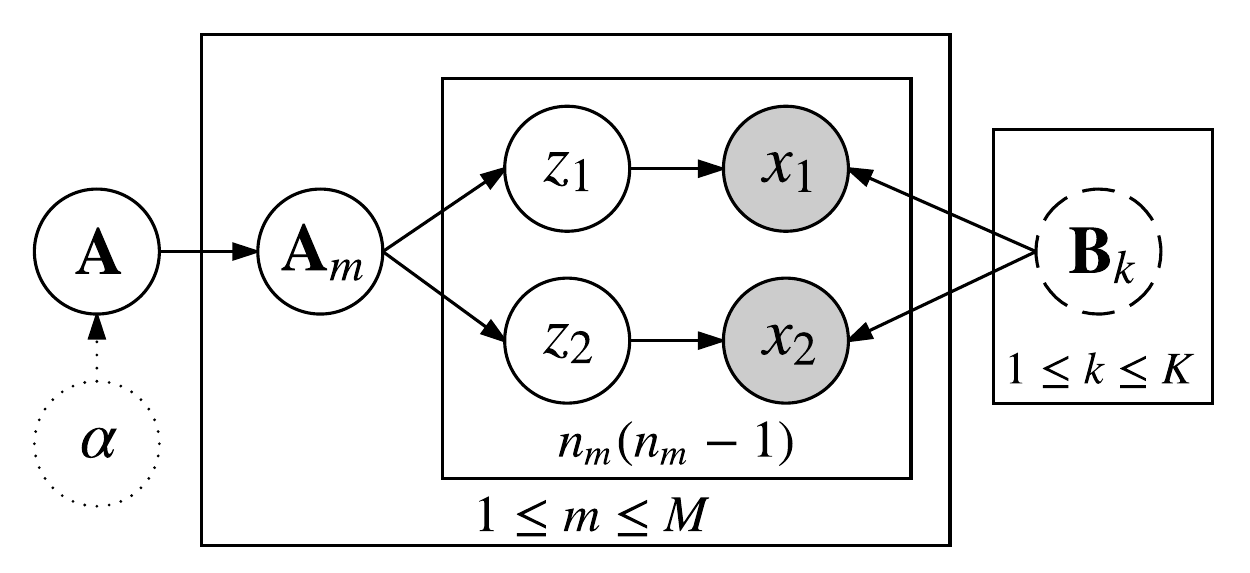}
		\caption {\small{JSMF asserts a {\em joint} distribution $\BA_m$ over topic pairs for each document $m$. $\BA$ serves as a prior for the entire corpus.}}
	\label{fig:JSMF}
	\vspace{-15px}
\end{figure}

In this section, we formalize matrix-based spectral topic modeling, especially JSMF.
We call particular attention to the presence of topic-topic matrix that represents joint distribution between pairs of topics in overall corpus.
This matrix will serve as a prior for document-specific joint probabilities between pairs of topics in later sections.

Suppose that a dataset has $M$ documents consisting of tokens drawn from a vocabulary of $N$ words.
Topic models assume that there are $K$ topics prepared for this dataset where each topic $k$ is a distribution $p(x | z=k)$ over $N$ words.
Denoting all topics by the column-stochastic matrix $\BB \In \R^{N \times K}$ where the $k$-th column vector $\Bb_k \in \Delta^{N-1}$ corresponds to the topic $k$, each document $m$ is written by first choosing a composition of topics $\Bw_m \In \Delta^{K-1}$ from a certain prior $\f$. 
Then from the first position to its length $n_m$, a topic $z$ is selected with respect to the composition $\Bw_m$, and a word $x$ is chosen with respect to the topic $\Bb_z$.
Different models adopt different $\f$.
For example, in the popular Latent Dirichlet Allocation (LDA), \cite{LDA} $\f = \Dir(\Balpha)$ as depicted in Figure \ref{fig:LDA}.
For the Correlated Topic Models (CTM), \cite{BL1}, $\f=\LN(\Bmu, \BSigma)$.

Let $\BH \In \R^{N \times M}$ be the word\hyp{}document matrix where the $m$-th column vector $\Bh_m$ indicates the observed term\hyp{}frequencies in the document $m$.
If we denote all topic compositions by another column\hyp{}stochastic matrix $\BW \In \R^{K \times M}$ whose $m$-th column vector is $\Bw_m \In \Delta^{K-1}$,
the two main tasks for topic modeling are to learn \textbf{topics} (i.e., the word-topic matrix $\BB$) and their \textbf{compositions} (i.e., the topic-document matrix $\BW$).
Inferring the latent variables $\BB$ and $\BW$ are coupled through the observed terms, making exact inference intractable.
Likelihood-based algorithms such as Variational EM and MCMC update both parts until convergence by iterating through documents multiple times.
If denoting the word-probability matrix by $\BHtilde$, which is the column-normalized $\BH$, these two learning tasks can be viewed as Non-negative Matrix Factorization (NMF): $\BHtilde \approx \BB \BW$, where $\BB$ and $\BW$ are also coupled.

\paragraph{Joint Stochastic Matrix Factorization}
The word-probability matrix $\BHtilde$ is highly noisy statistics due to its extreme sparsity, and it does not scale well with the size of dataset.
Instead, let $\BC \In \R^{N \times N}$ be the word co-occurrence matrix where $\BC_{ij}$ indicates the joint probability to observe a pair of words $(i, j)$.
Then we can represent the topic modeling as a second-order non-negative matrix factorization: $\BC \mathbin{\approx} \BB \BA \BB^T$ where we decompose the joint-stochastic $\BC$ into the column-stochastic $\BB$ (i.e., the word-topic matrix) and the joint-stochastic $\BA$ so called the topic-topic matrix.
If the ground-truth topic compositions $\BW^*$ that generate the data is known, we can define the posterior topic-topic matrix by $\BA^* \Ceq \frac{1}{M} \BW^* \BW^{*T} \In \R^{K \times K}$ where $\BA^*_{kl}$ indicates the joint posterior probability for a pair of latent topics $(k, l)$.
In this second-order factorization, $\BC$ is constructed as an unbiased estimator from which we can identify $\BB$ and $\BA$ close to the truthful topics and their correlations.\footnote{It is proven that the learned $\BA$ is close to the $\BA^*$ and the prior $\E_{\Bw \sim \f}[\Bw \Bw^T]$ (i.e., the population moment) for sufficiently large $M$. It allows us to perform topic modeling \cite{AGM}.}.

It is helpful to compare the matrix-based view of JSMF to the generative view of standard topic models.
The generative view focuses on how to produce streams of word tokens for each document, and the resulting correlations between words could be implied but not explicitly modeled.
In the matrix-based view, in contrast, we begin with word co\hyp{}occurrence matrix which explicitly models the correlations between words and produce \textbf{pairs of words} rather than individual words.
Given the prior topic correlations $\BA$ between pairs of topics, each document has its own \textbf{topic correlations} $\BA_m$ from $\BA$ as a joint distribution $p_m(z_1, z_2)$.\footnote{Strictly speaking, $\BA_m$ and $\BA$ (also $\BA^*$) are all joint distributions, neither covariances or correlations. However, as the covariance/correlations $\propto p(z_1, z_2) - p(z_1)p(z_2)$, which are directly inducible from $\BA's$, we keep using the naming convention from previous work, calling them \textit{topic correlations}.}
Then for each of the possible $n_m(n_m-1)$ pairs of positions, a topic pair $(z_1, z_2)$ is selected first from $\BA_m$, then a pair of words $(x_1, x_2)$ is chosen with respect to the topics ($\Bb_{z_1}, \Bb_{z_2})$, as illustrated in Figure \ref{fig:JSMF}.
Two important implications are:
\begin{itemize}
	\item The matrix of topic correlations $\BA$ represents the prior $\f$ without specifying any particular parametric family.
	\item $\BA_m$ is a rank-1 matrix $\Bw_m \Bw_m^T$ with $\Bw_m \sim \f$, providing the fully generative story for documents.
\end{itemize}

Note that the columns of $\BB$ in spectral topic models are sets of parameters rather than random variables which are sampled from another distribution $\g$ (e.g., $\g = \Dir(\Bbeta)$).
Other work relaxes this assumption \cite{nguyen2014anchors}, but we find that it is not an issue for the present work.
As putting a prior $\f$ over $\{\Bw_m\}$ is the crux of modern topic modeling \cite{asuncion09onsmoothing}, our flexible matrix prior $\BA$ allows us to identify the topics $\BB$ from $\BC$ without hurting the quality of topics.
However, learning $\BB$ and $\BA$ via the Anchor Word algorithms might seem loosely decoupled because the algorithms first recover $\BB$ and then $\BA$ from $\BB$ and $\BC$.
Previous work has found that rectifying $\BC$ is essential for quality spectral inference in JSMF \cite{moontae2015nips}.
The empirical $\BC$ must match the geometric structure of its posterior $\BB \BA^* \BB^T$, otherwise the model will fit noise.
Because this rectification step alternatingly projects $\BC$ based on the geometric structures of $\BB$ and $\BA$ until convergence, the rest of inference would no longer require mutual updates.

\paragraph{Related work}

Second-order word co-occurrence is not by itself sufficient to identify topics \cite{anandkumar2013}, so much work on second-order topic models adopts the \textit{separability assumption} such that each topic has an \textit{anchor word} which occurs only in the context of that topic.\footnote{Indeed, according to \cite{ding2015}, most large topic models are proven separable.}
However, the first Anchor Word algorithm \cite{AGM} is not able to produce meaningful topics due to numerical instability.
A second version \cite{arora2013practical} works if $K$ is sufficiently large, but the quality of topics is not fully satisfactory in real data even with the large enough $K$.
Also, this version is not able to learn meaningful topic correlations.
\cite{moontae2015nips} proposes the rectification step within JSMF, finally learning both quality topics and their quality correlations comparable to probabilistic inference in any condition.

There have been several extensions to the anchor words assumption that also provide identifiability.
The Catchwords algorithm \cite{bansal2014} assumes that each topic has a group of \textit{catchwords} that occur strictly more frequently in that topic than in other topics.
Their Thresholded SVD algorithm learns better topics under this assumption, but at the cost of slower inference.
Another minimal condition for identifiability is \textit{sufficient scatteredness} \cite{huang2016}.
The authors try to minimize the NMF objectives with an additional non-convex constraint, outperforming the second version of the Anchor Word algorithm \cite{arora2013practical} on smaller number of topics.

Another approach to guarantee identifiability is to leverage third-order moments.
The popular CP-decomposition \cite{hitchcock927} transforms the third-order tensor into a orthogonally decomposable form\footnote{This step is called the \textit{whitening}, which is \textit{conceptually} similar to the rectification in JSMF.} and learns the topics under the assumption that the topics are uncorrelated \cite{anandkumar2012}.
Another method is to perform Tucker decomposition \cite{Tucker}, which does not assume uncorrelated topics.
This approach requires additional sparsity constraints for identifiability and includes more parameters to learn \cite{anandkumar2013}.
While correlations between topics are not an immediate by-product of tensor-based models, the PADD method presented here is still applicable for learning topic compositions of these models if the modeler chooses proper priors that can capture rich correlations \cite{arabshahi2016spectral}.\footnote{One can also use the simple Dirichlet prior, although in theory it only captures negative correlations between topics.}

\section{Document-specific Topic Inference}
In Bayesian settings, learning topic compositions $\BW$ of individual documents is an inference problem which is coupled with learning topics $\BB$.
As each update depends also on the parametric prior $\f$ and its hyper-parameters $\Balpha$, $\Balpha$ must be optimized as well for fully maximizing the likelihood of the data \cite{wallach2009}. 
From higher-order moments, contrastively in spectral settings, we can recover the latent topics $\BB$ and their correlations $\BA$, the flexible prior that Bayesian models should painfully learn.\footnote{It means that the learned $\BA$ implies the information of the proper prior $\f(\Balpha)$ with respect to the data. If $\f=\Dir(\Balpha)$, we can indeed estimate $\Balpha$ via moment-matching $\E_{\Bw\sim\f}[\Bw\Bw^T]$ with $\BB$.} 
Since $\BB$ and $\BA$ are both provided and fixed, it is natural to formulate learning each column of $\BW$ as an estimation than an inference.

Besides revisiting the likelihood-based inferences, Thresholded Linear Inverse (TLI) is the only existing algorithm to the best of our knowledge, which is recently designed for the second-order spectral inference \cite{arorab16}. 
In this section, we begin with describing TLI and our SPI that only use the learned topics $\BB$, and then we propose our main algorithm PADD that uses the learned correlations $\BA$ as well.
By formulating the estimation as a dual decomposition \cite{komodakis2011mrf,rush2012tutorial}, PADD can effectively learn the compositions $\BW$ given $\BB$ and $\BA$.

\subsection{Simple Probabilistic Inverse (SPI)} 
Recall that selecting $n_m$ words in the document $m$ is the series of multinomial choices (i.e., $\Bh_m \sim \Mult(n_m, \BB \Bw_m)$). 
Denote $\Bh_m / n_m$ by $\Bhtilde_m$, then the conditional expectation satisfies $\E_{\Bw_m}[\Bhtilde_m] = \BB \Bw_m$.
If there is a left-inverse $\BBinv$ of $\BB$ that satisfies $\BBinv \BB \approx \BI_K$, then $\E_{\Bw_m}[\BBinv \Bhtilde_m] = \BBinv \BB \Bw_m \approx \Bw_m$.
However, not every left inverse is equivalent.
The less $\BBinv \BB$ is close to $I_K$, the more bias the estimation causes.
On the other hand, large entries of $\BBinv$ increases variance of the estimation.
Quality document-topic compositions could be recovered by finding one of the left-inverses that properly balances the bias and the variance.
One can choose $\BBinv$ as the optimizer that minimizes its largest entry $|\BBinv|_{\infty}$ under the small bias constraint: $|\BBinv \BB - \BI_K|_{\infty} \leq \delta$.

Let $\Bw_m^* \in \Delta^{K-1}$ be the true topic distribution that is used for generating the document $m$. 
Denoting the value $|\BBinv|_{\infty}$ at the optimum by $\lambda_{\delta}(\BB)$, one can bound the maximum violation $\|\BBinv \Bhtilde_m - \Bw_m^*\|_{\infty}$ by $ \delta + 2\lambda_{\delta}(\BB)\sqrt{(\log K)/n_m}$ for an arbitrary prior $\f$ from which $\Bw_m^* \sim \f$ \cite{arorab16}.
Thus the TLI algorithm first computes the best left-inverse $\BBinv$ of $\BB$ given the fixed $\delta$ and linearly predicts $\BW=\BBinv \BHtilde$ via one single estimator $\BBinv$.
Then for every column $\Bw_m$ of $\BW$, it thresholds out each of  the unlikely topics whose mass is smaller than $\tau \Eq 2\lambda_{\delta}(\BB)\sqrt{(\log K)/n_m} \Plus \delta$.
While TLI is supported by provable guarantees, it quickly loses accuracy if the given document $m$ exhibits correlated topics, its length $n_m$ is not sufficiently large, or $\Bw_m^*$ is not sparse enough.
In addition, since the algorithm does not provide any guidance on the optimal bias/variance trade-off, users might end up computing many inverses with different $\delta$'s.\footnote{Recall that computing this inverse is expansive and unstable.}

We instead propose the Simple Probabilistic Inverse (SPI) method, which is a one shot benchmark algorithm that predicts $\BW$ as $\BBreve \BHtilde$ without any additional learning costs.
Recall that Anchor Word algorithms first recover $\BBreve$ whose $\BBreve_{ki} \Eq p(z \Eq k | x \Eq i)$, and then convert it into $\BB$ whose $\BB_{ik} \Eq p(x \Eq i|z \Eq k)$ via Bayes rule \cite{arora2013practical}. 
For the probabilistic perspective, $\BBreve$ is a more natural linear estimator without having any negative entries like $\BBinv$.\footnote{Due to the low-bias constraint, $\BBinv$ is destined to have many negative entries, thus yielding negative probability masses on the predicted topic compositions $\BBinv \BHtilde$ even if its  pure column sums are all close to 1. While such negative masses are fixed via the thresholding step, equally zeroing-out both tiny positive masses and non-negligible negative masses is questionable.}
By construction, in contrast, the predicted topic composition via SPI is more likely to contain all possible topics that each word in the given document can be sampled from, no matter how negligible they are. 
But it can be still useful for certain applications that require extremely fast estimations with high recalls.
We later see in which conditions the SPI works reasonably well through the various experiments.

 \begin{figure}[t]
        \vspace{-10px}
		\centering
		\begin{algorithm}[H]
			\caption{Estimate the best compositions $\BW$. \\ (Master problem governing the overall estimation)}
			\textbf{def} \textsc{PADD}$(\BH, \BB, \BA, \lambda, \gamma)$\\\vspace{-10px}    
			\begin{algorithmic}[1]    
				\STATE $\BHtilde \gets $ column-normalize($\BH$)
				\STATE $\BLambda^{(0)} \gets \0^{K \times K}$, $\;\; \BW^0 \gets \BBreve \BHtilde$, $\;\; \BF \gets \gamma \BB^T \BHtilde$
				\REPEAT
					\STATE $\BG^{(t)} \gets (\gamma(\BB^T \BB + \frac{1}{M}\BLambda^{(t-1)}) + \BI_K)^{-1}$
					\FOR {\textbf{each} $m \in \{1, ..., M\} \; \text{(in parallel)} \;$}  
						\STATE $\Bf_m \gets \BF_m$
						\STATE $\Bw_m^0 \gets \BW_m^0$ (initial guess)
						\STATE $\Bwbar_m \gets \text{ADMM-DR}(\BG^{(t)}, \Bf_m, \Bw_m^0, \lambda)$
				\ENDFOR\\
				\STATE $\BLambda^{(t)} \gets \BLambda^{(t-1)} - \tau_t (\BA - \frac{1}{M}\sum_{m=1}^M (\Bwbar_m \Bwbar_m^T)))$
				\UNTIL $\text{the convergence}$\\
				\RETURN $\BW = \{\Bwbar_1 | ... | \Bwbar_M\}$
			\end{algorithmic}
			\label{alg:MASTER}
		\end{algorithm}
		\vspace{-15px}
\end{figure}

\subsection{Prior-aware Dual Decomposition (PADD)}
To better infer topic compositions, PADD uses the learned correlations $\BA$ as well as the learned topics $\BB$. 
While people have been more interested in finding better inference methods, many algorithms including the family of Variational Bayes and Gibbs Sampling turn out to be different only in the amount of smoothing the document-specific parameters for each update \cite{asuncion09onsmoothing}.
On the other hand. a good prior $\f$ and the proper hyper-parameter $\Balpha$ are critical, allowing us to perform successful topic modeling with less information about documents, but their choices are rarely considered \cite{wallach2009}.

Second-order spectral models do not specify $\f$ as a parametric family $\f(\Balpha)$, but the posterior topic-topic matrix $\BA^*$ closely captures topic prevalence and correlations.
Since the learned $\BA$ is close to $\BA^*$ given a sufficient number of documents, one can estimate better topic compositions by matching the overall topic correlations (by $\BA$) as well as the individual word observations (by $\BB$).\footnote{Because the learned $\BB$ and the posterior moment $\BA^*$ are close to the population moment $\E_{\Bw \sim \f}[\Bw \Bw^T]$ if $M$ is sufficiently large, PADD might not able to find quality compositions if both $M$ and $n_m$ are small. However, this problem also happens in probabilistic topic models, which is due to lack of information.}
For a collection of $M$ documents, PADD tries to find the best compositions $\BW = \{\Bw_m\}$ that satisfies the following optimization:
\begin{align}
\label{eqn:PADD}
\min \quad& \sum_{m=1}^M \| \BB \Bw_m - \Bhtilde_m\|_2^2 \\
\text{subject to} \quad& \Bw_m \In \Delta^{K-1} \;\; \text{and}  \;\; \frac{1}{M}\sum_{m=1}^M \Bw_m \Bw_m^T = \BA. \nonumber
\end{align}
Solutions from \eqref{eqn:PADD} tries to match the observed word-probability $\Bhtilde_m$ as individuals (i.e., loss minimization), while simultaneously matching the learned topic correlations $\BA$ as a whole (i.e., regularization).
Therefore, whereas the performance of TLI depends only on the quality of the estimated word-topic matrix $\BB$, PADD also leverages the learned correlations $\BA$ to perform an analogous estimation to the prior-based probabilistic inference.
With further tuning of the balance between the loss and the regularization with respect to the particular task, PADD can be more flexible for various types of data, whose topics might not empirically well fit to any known parametric prior.

\begin{figure}[t]
        \vspace{-10px}
		\centering
		\begin{algorithm}[H]
			\caption{Estimate the best individual $\Bw_m$. \\ (Subproblem running for each document $m$ in parallel)}
			\textbf{def} \textsc{ADMM-DR}$(\BG, \Bf, \Bw^{(0)}, \lambda)$\\\vspace{-10px}
			\begin{algorithmic}[1]        			
				\STATE $\Bq^{(0)} \gets \Bw^{(0)}$
				\REPEAT 
					\STATE $\Bp^{(t)} \gets \BG(2\Bw^{(t-1)} - \Bq^{(t-1)} + \Bf)$     
					\STATE $\Bq^{(t)} \gets \Bq^{(t-1)} + \lambda(\Bp^{(t)} - \Bw^{(t-1)})$
					\STATE $\Bw^{(t)} \gets \Pi_{\Delta^{K-1}}(\Bq^{(t)})$
				\UNTIL $\text{the convergence of} \;\; \Bw^{(t)}$\\   
				\STATE $\Bwbar \gets \Bw^{(t)}$
				\RETURN {$\Bwbar$}    
			\end{algorithmic}
			\label{alg:SUB}
			\vspace*{-1px}
			\hrulefill\\
			($\Pi_{\Delta^{K-1}}(\cdot)$ is the orthogonal projection to the simplex $\Delta^{K-1}$. See the reference for the detailed implementation \cite{duchi2008efficient}.)  
		\end{algorithm}	
		\vspace{-15px}
\end{figure}

\begin{figure*}[t]
 	\centering
	\includegraphics[width=0.95\textwidth, trim={1.0cm 1.0cm 1.0cm 0.0cm}]{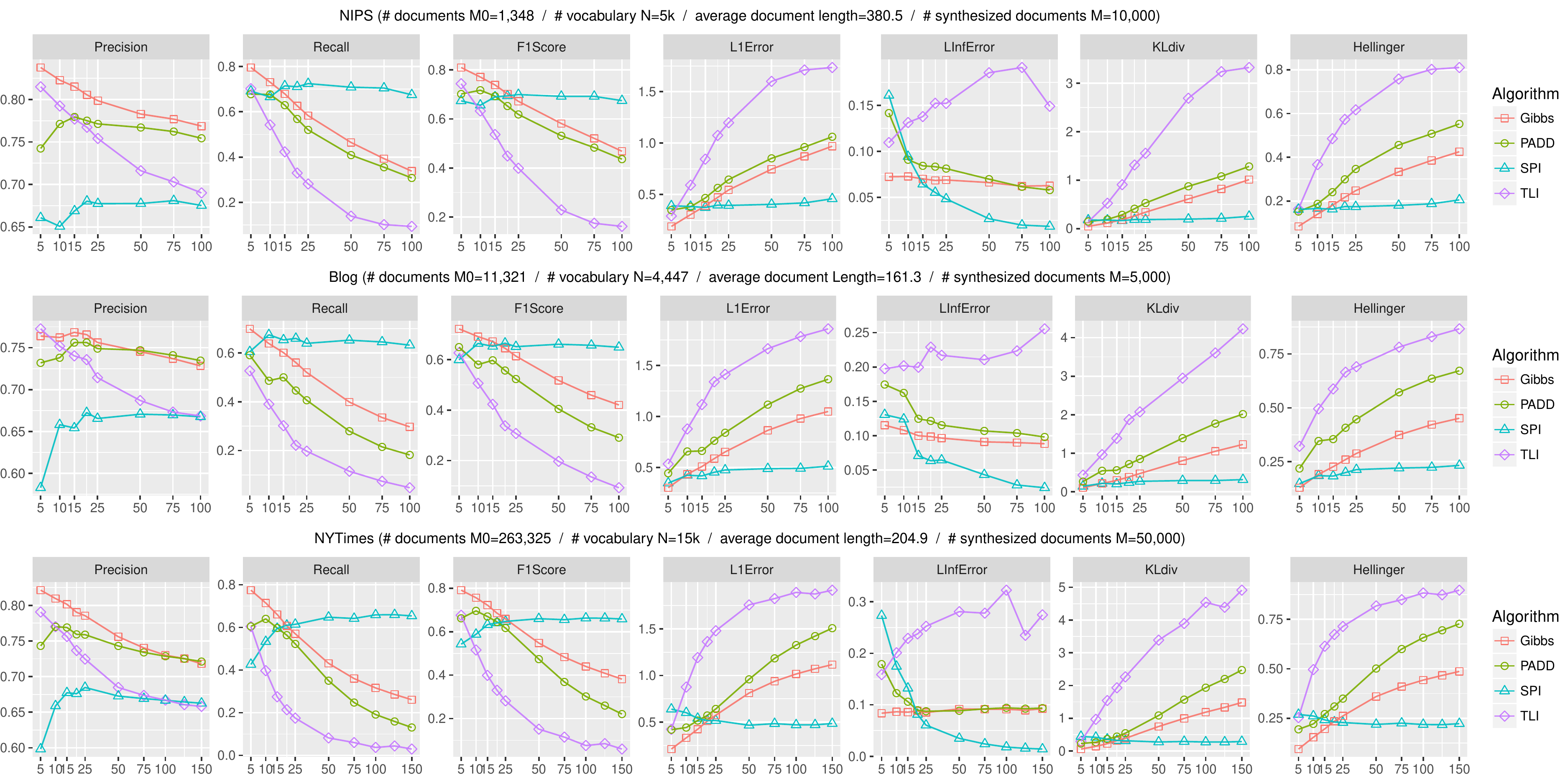}
	\caption {\small{Artificial experiment on Semi-Synthetic (\textit{SS}) corpus with highly sparse topics but little correlations. X-axis: \#  topics $K$. Y-axis: higher numbers are better for the left three columns, lower numbers are better for the right four. SPI performs the best with $K \geq 25.$}}
	\label{fig:semisynth}		
\end{figure*}

\subsection{Parallel formulation with ADMM}
It is not easy to solve \eqref{eqn:PADD} due to the non-linear coupling constraint $(1/M)\sum \Bw_m \Bw_m^T \Eq \BA$.
We can construct a Lagrangian by adding a symmetric matrix of dual variables $\BLambda \In \R^{K \times K}$. 
Then $\Lagr(\Bw_1, ..., \Bw_M, \BLambda)$ is equal to
\begin{align}
	 &\sum_{m=1}^M \| \BB \Bw_m - \Bhtilde_m \|_2^2 \;\; + \langle \BLambda, \bigg(\frac{1}{M}\sum_{m=1}^M \Bw_m \Bw_m^T \bigg) \Minus \BA \rangle_F \nonumber \\
	   &\Eq \sum_{m=1}^M \bigg\{\| \BB \Bw_m \Minus \Bhtilde_m \|_2^2 \Plus \frac{1}{M} \langle \BLambda, \Bw_m \Bw_m^T \Minus \BA \rangle_F \bigg\} \hspace{-.5em}
	\label{eqn:DD}
\end{align}
The equation \eqref{eqn:DD} implies that given a fixed dual matrix $\BLambda$, minimizing the Lagrangian can be decomposed into $M$ subproblems, allowing us to use the dual decomposition \cite{komodakis2011mrf,rush2012tutorial}.
Each subproblem tries to find the best topic composition $\Bw_m \In \Delta^{K-1}$ that minimizes $ \| \BB \Bw_m \Minus \Bhtilde_m \|_2^2 \Plus (1/M) \langle \BLambda, \Bw_m \Bw_m^T \Minus \BA \rangle_F$.\footnote{The operation $\langle\cdot, \cdot\rangle_F$ indicates the Frobenius product, which is the matrix version of the inner product.}
Once every subproblem is solved and has provided the current optimal solution $\Bwbar_m$, the master problem simply updates the dual matrix based on its subgradient: $\smash{-\frac{1}{M}(\sum_{m=1}^M (\Bwbar_m \Bwbar_m^T - \BA)) \In \partial(\BLambda)}$, and then distributes it back to each subproblem.
For robust estimation, we adopt the Alternating Direction Method of Multiplier (ADMM) \cite{bioucas2010alternating} with Douglas-Rachford (DR) splitting \cite{li2016douglas}.
Then the overall procedures become as illustrated in Algorithm \ref{alg:MASTER} and \ref{alg:SUB}.

Note first that master problem in Algorithm \ref{alg:MASTER} computes the matrix inverse, but the computation is cheap and stable. 
This is because it only algebraically inverts $K \Times K$ matrix rather than solving a constraint optimization to invert $N \Times K$ matrix as TLI does.
Note also that the overall algorithm repeats the master problem only small number of times, whereas each subproblem repeats the convergence loop more times.
The exponentiated gradient algorithm \cite{arora2013practical} is also applicable for quick inference, but tuning the learning rate would be less intuitive, although PADD should be also careful in tuning the learning rate $\tau_t$ due to its non-linear characteristics.
Note last that we need not further project the subgradient to the set of symmetric matrices because only the symmetric matrices $\BA$ and $\Bwbar_m \Bwbar_m^T$ are added and subtracted from $\BLambda$ for every iteration of the master problem.

\paragraph{Why does it work?} 
Probabilistic topic models try to infer both topics $\BB$ and document compositions $\BW$ that approximately maximize the marginal likelihood of the observed documents: $\prod_{m=1}^M \int_{\Bw_m} p(\Bw_m | \Balpha)\prod_{i=1}^N (\BB \Bw_m)_i^{\Bh_{mi}} d\Bw_m$, considering all possible topic distributions $\Bw_m \In \Delta^{K-1}$ under the prior $\f(\Balpha)$.
However, if the goal in spectral settings is to find the best individual composition $\Bw_m$ provided with the learned $\BB$ and $\BA$, the following MAP estimation 
\begin{equation}
    \argmax_{\Bw_m \in \Delta^{K-1}} p(\Bw_m ;\BA)\prod_{i=1}^N (\BB \Bw_m)_i^{\Bh_{mi}} \;\; (\Bh_{mi} = \BH_{im}) \label{eqn:likelihood}
\end{equation} 
is a reasonable pointwise choice in the likelihood perspective.
Recall that the MLE parameters that maximize the likelihood of the multinomial choices $\Bh_m$ is to assign the word-probability parameters $\BB\Bw_m$ equal to the empirical frequencies $\Bhtilde_m$. 
The loss function of the PADD objective tries to find the best $\Bw_m$ that makes $\BB \Bw_m \approx \Bhtilde_m$, maximizing the second term $\prod_{i=1}^N (\BB\Bw_m)_i^{\Bh_{mi}}$ in Equation \eqref{eqn:likelihood}.

While sampling a rank-1 correlation matrix $\BA_m \Eq \Bw_m \Bw_m^T$ from $\BA$ is not yet known but an interesting research topic, it is true that PADD tries to maximize the first term $p(\Bw_m ; \BA)$ in \eqref{eqn:likelihood} by preventing $\Bw_m$'s far deviation from the learned topic correlations $\BA$, which is a good approximation of the prior: the population moment $\E_{\Bw \sim \f}[\Bw \Bw^T]$.
Indeed, when learning the document-specific topic distributions for spectral topic models, it is shown that a proper point estimation is likely a good solution also in the perspective of Bayesian inference because the posterior is concentrated on the $\epsilon$-ball of the point estimator with large chance \cite{arorab16}.

\section{Experimental Results}

\begin{figure*}[t]
 	\centering
	\includegraphics[width=0.95\textwidth, trim={1.0cm 1.0cm 1.0cm 0.0cm}]{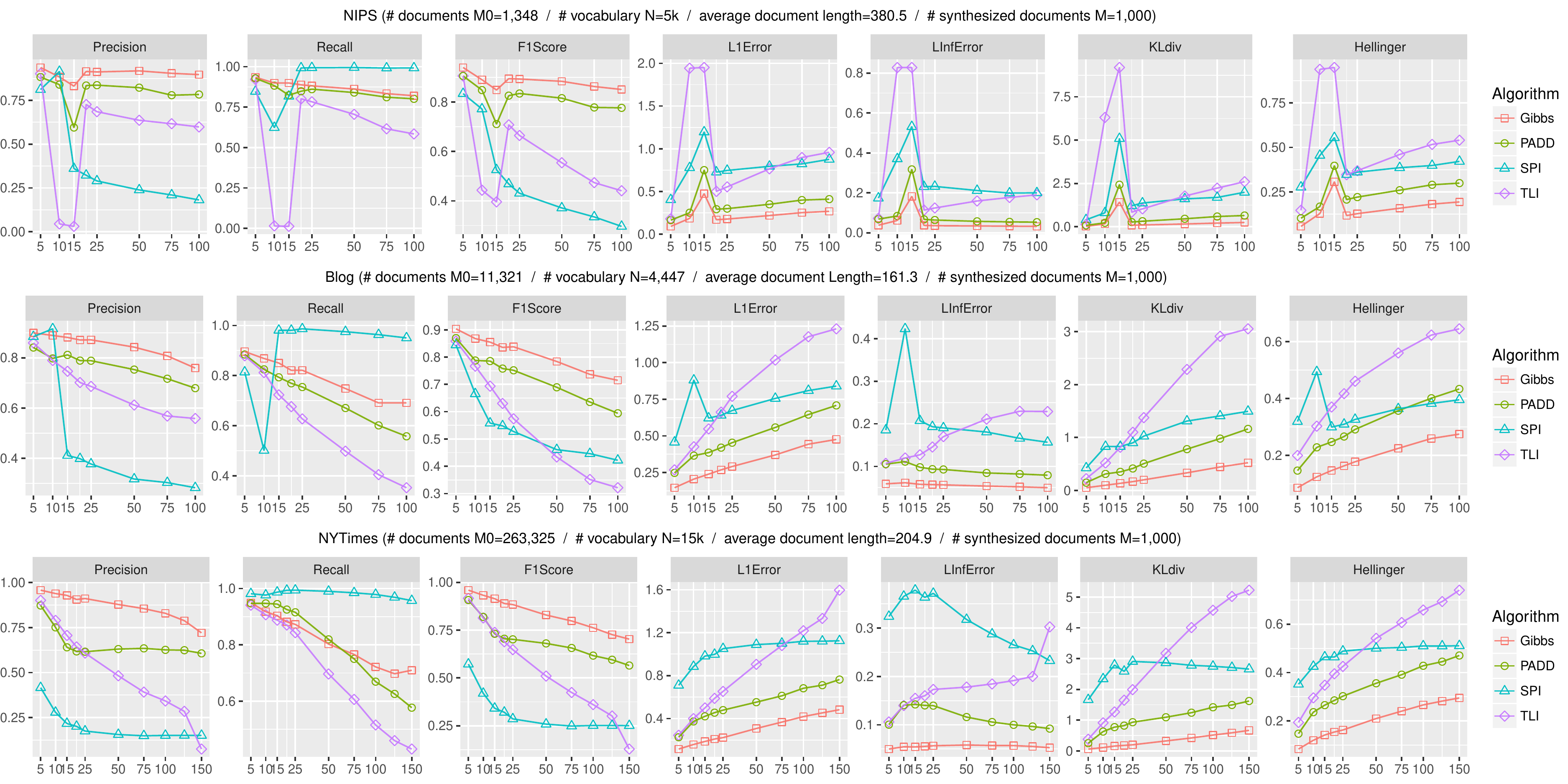}
	\caption {\small{Realistic experiment on Semi-Real (\textit{SR}) corpus with non-trivial topic correlations. X-axis: \# topics $K$. Y-axis: higher numbers are better for the left three columns, lower numbers are better for the right four. PADD is consistent and comparable to Gibbs Sampling.}}
	\label{fig:semireal}		
\end{figure*}

Evaluating the learned topic compositions is not easy for real data because no ground truth compositions exist for quantitative comparison.
In addition, qualitatively measuring coherency of topics in each document composition is no longer feasible because there are too many documents, and topics in each document may not be as obviously coherent as words in each topic \cite{chang2009reading}.
Synthesizing documents from scratch is an option as we can manipulate the ground truth $\BB$ and $\BW^*$, but the resulting documents would not be realistic.
Thus we use two generative processes to create two classes of semi documents that emulate the properties of real documents but that are also by construction compatible with specific models.

The semi-synthetic setting involves sampling from a fitted LDA based on Dirichlet prior, whereas the semi-real setting involves sampling from a fitted CTM \cite{BL1} based on Logistic-normal prior.
Given the original training data $\BH_0$ and the number of topics $K$, we first run JSMF$(\BC_0, K)$\footnote{$\BC_0$ is constructed form $\BH_0$ as an unbiased estimator.} \cite{moontae2015nips} and CTM-Gibbs($\BH_0$, K) \cite{chen2013scalable}, learning $(\BB_0, \BA_0, \BBreve_0)$ and $(\Bmu_0, \BSigma_0, \BB_0)$, respectively.
For each number of documents $M$, we synthesize Semi-Synthetic (\textit{SS}) corpus $\BH_{SS}$ and Semi-Real (\textit{SR}) corpus $\BH_{SR}$ by first sampling each column of the ground-truth topic compositions $\BW^*$ from $\Dir(\Balpha)$ and $\LN(\Bmu_0, \BSigma_0)$, respectively, and then sample each document $m$ with respect to $\BB_0$ and $\Bw_m^*$.\footnote{We randomly synthesize \textit{SS} and \textit{SR} so that their average document lengths coincide with its original training corpus $\BH_0$.}
While it is less realistic, \textit{SS} uses the sparsity controlled parameter $\Balpha \Eq (5/K)\vec{1}$ for fair comparison to the experiemtns of TLI in \cite{arorab16}, whereas \textit{SR} exploits the learned hyper-parameter $(\Bmu_0, \BSigma_0)$ so that it can maximally simulate the real world characteristics with non-trival correlations between topics
For Fully-Real (\textit{FR}) experiment, we also prepare the unseen documents $\BH_{US}$, which is 10\% of the original data which has never been used in training (i.e., $\BH_0 \cap \BH_{US} \Eq \emptyset$).
In \textit{FR}, we also test on the training set $\BH_0$.

As strong baselines, we run TLI and SPI on the \textit{SS} corpus $\BH_{SS}$ with the ground-truth real parameters $\BB_0$ and $\BBreve_0$, while Gibbs Sampling uses both $\BB_0$ and the ground-truth hyper-parameter $(5/K)\vec{1}$.
For \textit{SR} corpus, we run both TLI and SPI as isomorphic to \textit{SS} with $\BH_{SR}$.
Since we only know the Logistic-Normal parameters for \textit{SR}, however, Gibbs Sampling uses the approximated the Dirichlet hyper-parameter $\Balpha$ via matching the topic-topic moments between $(1/M)\BW^*\BW^*$ and $\E_{\Bw \sim \Dir(\Balpha)}[\Bw \Bw^T]$.\footnote{It is done by solving  the over-determined system. We verified that our learned hyper-parameter $\Balpha$ outperforms exhaustive line-search in terms of the moment-matching loss.}
In contrast, we run PADD with both the learned topics $\BB_0$ and their correlations $\BA_0$ for each of $\BH_{SS}$ and $\BH_{SR}$.
While the goal of synthetic experiments is to compare the learned $\BW$ to the ground-truth $\BW^*$, we cannot access $\BW^*$ for real experiments.
For \textit{FR} corpora $\BH_0$ and $\BH_{US}$, therefore, Gibbs Sampling with $\BB_0$ and the fitted $\Balpha$ (again by moment-matching based on $\BA_0$) serves as the ground-truth $\BW^*$.
We then run TLI, SPI, and PADD similarly with $\BB_0$, $\BBreve_0$, and $\BA_0$.

\begin{table*}[t]
\centering
\begin{tabular}{c|c|ccc|ccc|cc}
\hline
Data & Algo & Precision & Recall & F1-score & $\ell_1$-error & $\ell_\infty$-error & Hellinger & Prior-dist & Non-supp\\
\hline
NIPS & Rand & .092/.107 & .426/.429 & .260/.268 & 1.60/1.55 & 0.52/0.44 & 0.74/0.72 & .0123/.0126 & .908/.894 \\
(1.3k/ & TLI & .396/.421 & .954/.890 & .513/.536 & 0.88/0.88 & 0.29/0.24 & 0.44/0.45 & .0085/.0080 & .617/.604 \\
143) & SPI & .169/.193 & .994/.994 & .271/.307 & 1.20/1.14 & 0.41/0.34 & 0.55/0.53 & .0118/.0118 & .771/.755 \\
 & \textbf{PADD} & \textbf{.691}/\textbf{.734} & \textbf{.876}/\textbf{.740} & \textbf{.759}/\textbf{.744} & \textbf{0.47}/\textbf{0.55} & \textbf{0.16}/\textbf{0.18} & \textbf{0.27}/\textbf{\textbf{0.31}} & \textbf{.0049}/\textbf{.0016} & \textbf{.382}/\textbf{.324}\\\hline
Blog & Rand & .095/.100 & .424/.428 & .261/.263 & 1.55/1.53 & 0.49/0.46 & 0.70/0.69 & .0101/.0100 & .905/.900 \\
 (11k/ & TLI & .421/.422 & .856/.814 & .530/.527 & 0.95/0.97 & 0.28/0.27 & 0.47/0.49 & .0066/.0067 & .619/.623 \\
 1.2k) & SPI & .170/.176 & .982/.979 & .279/.289 & 1.18/1.16 & 0.39/0.37 & 0.51/0.50 & .0106/.0107 & .782/.776 \\
 & \textbf{PADD} & \textbf{.642}/\textbf{.671} & \textbf{.800}/\textbf{.758} & \textbf{.713}/\textbf{.715} & \textbf{0.61}/\textbf{0.62} & \textbf{0.21}/\textbf{0.21} & \textbf{0.33}/\textbf{\textbf{0.34}} & \textbf{.0057}/\textbf{.0044} & \textbf{.428}/\textbf{.403}\\\hline
NY & Rand & .061/.062 & .427/.427 & .207/.208 & 1.73/1.73 & 0.62/0.61 & 0.80/0.79 & .0188/.0188 & .939/.938 \\
Times & TLI & .321/.320 & .925/.922 & .428/.427 & 1.16/1.16 & 0.40/0.40 & 0.56/0.56 & .0078/.0197 & .690/.690 \\
(263k/ & SPI & .183/.183 & .964/.964 & .281/.281 & 1.29/1.29 & 0.47/0.47 & 0.58/0.58 & .0143/.0143 & .775/.775 \\
30k) & \textbf{PADD} & \textbf{.474}/\textbf{.543} & \textbf{.899}/\textbf{.904} & \textbf{.575}/\textbf{.668} & \textbf{0.76}/\textbf{0.71} & \textbf{0.27}/\textbf{0.28} & \textbf{0.39}/\textbf{\textbf{0.37}} & \textbf{.0005}/\textbf{.0113} & \textbf{.520}/\textbf{.493}\\\hline
\end{tabular}
	\caption {\small{Real experiment on Fully-Real (\textit{FR}) corpora. For each entry, a pair of values indicates the corresponding metrics on training/unseen documents. Averaged across all models with different $K$'s. Rand estimates randomly. For two new metrics: Prior-dist and Non-supp, smaller numbers are better. PADD performs the best considering topic compositions learned by Gibbs Sampling as the ground-truth.}}
\label{tab:fullyreal}
\end{table*}

To support comparisons with previous work, we use the same standard datasets: NIPS papers and NYTimes articles with the same vocabulary curation as \cite{moontae2015nips}.
By adding political blogs \cite{eisenstein2010cmu}, the sizes of training documents in our datasets exhibit various orders of magnitudes.
For evaluating information retrieval performance, we first find the prominent topics whose cumulative mass is close to 0.8 for each document, and compute the precision, recall, and F1 score as \cite{yao2009efficient}.
For measuring distributional similarity, we use KL-divergence and Hellinger distance.
In contrast to assymetric KL, Hellinger is a symmetric and normalized distance used in evaluating the CTM.
For comparing the reconstruction errors with TLI, we also report $\ell_1$-error and $\ell_\infty$-error \cite{arorab16}.
For fully real experiments, we additionally report the distance to prior $\|\BA_0 - (1/M)\BW\BW^T\|_F$ and the mass on non-supports, the total probability that each algorithm puts on non-prominent topics of the ground-truth composition.

In the Semi-Synthetic (\textit{SS}) experiments given in Figure \ref{fig:semisynth}, SPI performs the best as the number of topics $K$ increases.
As expected, SPI is good at recalling all possible topics though loosing precision at the same time.
Note that Gibbs Sampling shows relatively high $\ell_1$-error especially for the models with large $K$. 
This is because $\Dir((5/K)\vec{1})$ puts too sparse prior on topics, causing small miss predictions to be notably harmful even with sufficiently mixed Gibbs Sampling.\footnote{We throw out the first 200 burn-in iterations and keep going 1,000 further iterations to gather the samples from the posterior. We use the JAVA version of Mallet but providing the learned topics $\BB_0$ and the fitted hyper-parameter based on $\BA_0$ as fixed information. Only the topic composition $\BW$ is updated over iterations.}
The same problem happens also in \cite{arorab16}.
Despite the unrealistic nature of $\textit{SS}$, PADD outperforms TLI by a large margin in most cases, showing similar behaviors to probabilistic Gibbs Sampling across all datasets.
TLI performs well only for tiny topic models.

The situation is quite different in Semi-Real (\textit{SR}) experiments shown in Figure \ref{fig:semireal}. 
High recalls of SPI is no longer helpful due to the drastic loss of precision, whereas PADD is comparable to Gibbs Sampling across all models and datasets even with only 1k documents.
This is because PADD captures the rich correlations through its prior-aware formulation.
TLI performs poorly due to its linear nature without considering topic correlations.
When testing on Fully-Real (\textit{FR}) corpora, PADD shows the best performance on both training documents and the unseen documents. 
Considering Gibbs Sampling with the ground-truth parameters still loses the accuracy in other settings, the metrics evaluated against the Gibbs Sampling in Table \ref{tab:fullyreal} is noteworthy.
Prior-dist, the Frobenius distance to the prior $\BA_0$, implies PADD-learned $\Bw_m$ likely improves $p(\Bw_m ; \BA_0)$ than other algorithms.
While TLI uses provably guaranteed thresholding,\footnote{We use the same less conservative threshold $\tau/4.5$ and unbias setting with $\delta=0$ as conducted in \cite{arorab16}. We also try to loosen the unbias constraint when the inversion is failed. However it does not help.} Non-supp values show that it still puts more than half of probability mass on the non-prominent topics in average.

For inferring topic compositions by PADD, we iterates 15 times for the master procedure PADD and 150 times for the slave procedure ADMM-DR with $(\lambda, \gamma)\Eq(1.9, 3.0)$.
We verify that PADD converges well despite its non-convex nature. 
Inference quality is almost equivalent when running only 100 times for each slave procedure, and is not sensitive to these parameters as well.
When using the equivalent level of parallel processing, computing $\BBinv_0$ via TLI takes 2,297 seconds,\footnote{We also observe that AP-rectification in JSMF significantly boosts the condition of $\BB_0$ on various datasets, removing TLI's failures in computing the left-inverse $\smash{\BBinv_0}$. However, even if the inverse is computed, TLI sometimes yields NaN values due to numerical instability of matrix inversion. We omit those results in evaluation to prevent TLI's quality degradation.} whereas PADD takes 849 seconds for the entire inference on the semi-synthetic NIPS dataset with $K \Eq 100$ and $M \Eq 10,000$. 
SPI is the by far fastest without requiring anything than one matrix multiplication, whereas Gibbs takes 3,794 seconds on the same machine.
While we choose ADMM-DR mainly for the tightest optimization, one can easily incorporate faster gradient-based algorithms inside our formulation of prior-aware dual decomposition.

\section{Discussion}
\label{sec:Discussion}

Fast and accurate topic inference for new documents is a vital component of a topic-based workflow, especially for spectral algorithms that do not by themselves produce document-topic distributions.
We first verify that our Simple Probabilistic Inverse (SPI) performs well on semi-synthetic documents that have no meaningful correlation structure.
When a word is contributed to two or more topics based on the overall data, SPI naively distributes its prediction weights to those topics without considering the potential contribution of other co-occurring words and their correlations via underlying topics within the specific query document.
However one can also threshold or post-process SPI's estimation analogous to TLI, proposing a future research idea for extremely fast composition inference.

We then verify the power of Prior-aware Dual Decomposition (PADD).
Being armed with theoretical motivation and efficient parallel implementation, PADD performs comparable to probabilistic Gibbs Sampling especially for realistic data. 
The experimental results show that PADD also well predicts the topic compositions of unseen real data, notably outperforming the existing TLI method.
With robust and efficient topic inference that is aware of topical correlations latent in the data, we can now fill out the necessary tools to make spectral topic models a full competitor to likelihood-based methods.
Although the benefits of PADD for topic inference are mostly demonstrated in second-order spectral methods, but are applicable in any topic inference setting.



\bibliographystyle{icml2017}
\small{
	\bibliography{references}
}

\end{document}